# Deep Learning Based Early Diagnostics of Parkinson's Disease

Elcin Huseyn

Research Laboratory of Intelligent Control and Decision Making Systems in Industry and Economics, Azerbaijan State Oil and Industry University,
20 Azadlig Ave., Baku, AZ1010, Azerbaijan
elcin.huseyn@asoiu.edu.az

**Abstract.** In the world, about 7 to 10 million elderly people are suffering from Parkinson's Disease (PD) disease. Parkinson's disease is a common neurological degenerative disease, and its clinical characteristics are Tremors, rigidity, bradykinesia, and decreased autonomy. Its clinical manifestations are very similar to Multiple System Atrophy (MSA) disorders. Studies have shown that patients with Parkinson's disease often reach an irreparable situation when diagnosed, so As Parkinson's disease can be distinguished from MSA disease and get early diagnosis, people are constantly exploring new methods. With the advent of the era of big data, deep learning has made major breakthroughs in image recognition and classification. Therefore, this study proposes to use The deep learning method realizes the diagnosis of Parkinson's disease, multiple system atrophy and healthy people. This data source is from Istanbul University Cerrahpasa Faculty of Medicine Hospital. The processing of the original magnetic resonance image (Magnetic Resonance Image, MRI) is guided by the doctor of Istanbul University Cerrahpasa Faculty of Medicine Hospital. The focus of this experiment It is to improve the existing neural network so that it can obtain good results in medical image recognition and diagnosis. An improved algorithm was proposed based on the pathological characteristics of Parkinson's disease, and good experimental results were obtained by comparing indicators such as model loss and accuracy.

## 1. Introduction

### 1.1. Research Background

Parkinson's Disease (PD), also known as tremor paralysis, is the second most common degenerative disease of the central nervous system in the elderly. It is characterized by the gradual loss of dopaminergic neurons in the dense substantia nigra and impaired motor function [1], and It is characterized by the progressive loss of DopAminergic (DA) neurons. [2]. The average age of the disease is about 60 years old, and Parkinson's disease is less common in young people under 40. According to statistics, About 7 million to 10 million elderly people in the world are suffering from the disease. The prevalence of PD in people over 65 years of age in China is about 1.7% [3]. Most Parkinson's patients are sporadic cases, only Less than 10% of patients have a family history, so the etiology and pathogenesis of PD has not been clarified so far.

Multiple System Atrophy (MSA) is a slowly progressive neurodegenerative disease [4], which is characterized by MSA-P subtypes with hypokinesia and Parkinson's syndrome. Ataxia (cerebellar ataxi) dominates ataxia in the MSA-C subtype.

Magnetic Resonance Imaging (MRI) is a type of tomography, which uses magnetic resonance to obtain electromagnetic signals from the human body and reconstruct human information. At present, this method has been widely used in medical imaging. In the resonance image, we can obtain a variety of physical parameters of the substance, such as proton density, spin-lattice relaxation time T1, spin-spin relaxation time T2, diffusion coefficient, magnetization coefficient, chemical shift, etc. In this experiment, according to the doctor's recommendation, three types of MRI images: Diffusion Weighted Imaging (DWI), T2 and Coronal T2 Water Suppression Sequence (CorT2).

## 1.2 Research Status

At present, medical technology is developing rapidly, but most of the diagnosis of Parkinson's disease can be confirmed by clinical symptoms of patients; however, domestic and foreign experts have proved through a large number of clinical experience and experiments that the main pathological change of Parkinson's disease is the progressive nature of nigrostriatal cells Loss and accumulation of intracellular Lewis bodies, from degeneration of nigra and striatum DA neurons-loss of clinical symptoms requires a long preclinical process, with an incubation period of about 5 years, and niger DA neurons are lost <50 Clinical symptoms are not obvious in% of patients. When patients have clinical symptoms of PD, DA energy neurons in the brain are lost by 70% ~ 80% [5].

On the MR high-resolution T2WI weighted image / magnetically sensitive weighted imaging (SWI), the normal nigral nucleus-1 axis resembles a dovetail, called a dovetail sign. The pathological feature of PD is the progressive dopaminergic neurons present in the substantia nigra. The previous study found that there are 5 substantia nigras in the substantia nigra species, and the largest substantia nigra-1 is the structure that mainly affects the pathological changes of PD. Substantia nigra-1 is located in the rear 1/3 of the substantia nigra, and is axial. SWI appears as a bar-shaped or comma-shaped high signal, similar to a dovetail. The "swallowtail" represented by nigrosomes is surrounded by low SWI signals in the front, sides and inside, and bifurcations are visible. The "swallowtail sign" disappears for diagnosis The accuracy rate of Kingson's disease is about 90%. For patients with Parkinson's disease, the signal of nigrosomal-1 is low and the swallowtail sign disappears [6], as shown in Figure 1.

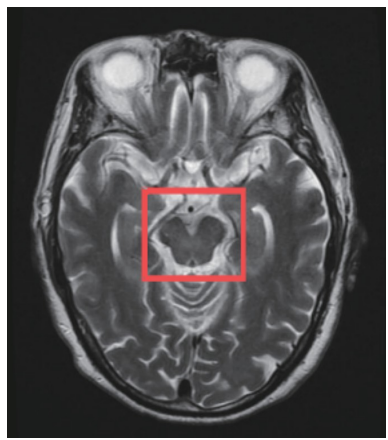

**Fig.1.** Brain map of PD patients (labeled with "Swallowtail" area in the center)

Routine MRI examinations can show: 1 Brain atrophy: mainly the extrapyramidal atrophy causes the third ventricle to widen, and the cerebral sulcus widens due to diffuse atrophy of the cerebral cortex. 2 Atrophy of the dense substantia nigra: weighted on T2 weighted images / proton density On the image, hesitation of the presence of high-concentration iron in the reticular zone and red nucleus of the nevus of normal brain tissue showed a low signal; the sound of the dense band with a low iron concentration showed a local signal, etc. In addition, PD patients can also be seen due to degeneration and necrosis of melanocytes. And iron metabolism has become a result of narrowing dense bands, blurred edges and other manifestations. By observing the morphology and signal changes of the substantia nigra, measuring the width of the substantia nigra and the ratio of the width of the substantia nigra to the midbrain, for the diagnosis of PD Differential diagnosis of PD and vascular Parkinson's syndrome provides an objective basis. 3 The striatum area on the T2-weighted image caused by low iron deposition in the posterolateral nucleus of the putamen shows a low signal [7].

So far, research that combines deep learning methods and Parkinson's disease diagnosis has mainly focused on the following aspects. Al-Fatlawi and Jabardi et al. Proposed the use of Deep Belief Network (DBN) [8] for Parkinson's disease diagnosis, of which The information analyzed is the patient's speech signal. The deep belief network is composed of two restricted Boltzmann machines [9] and an output layer. The first one performs unsupervised learning and the second one performs fine-tuning of back propagation. Supervised learning. The accuracy of the test reached 94% in this study. Shamir and Dolber [10] and others proposed the use of deep learning methods to detect the degree of limb retardation of patients and then perform classification diagnosis. In comparing traditional machine learning methods and convolutional nerve-based methods Among the deep learning methods of the network, deep learning is better than other machine learning methods by 4.6 percentage points in accuracy. The above studies are based on the combination of deep learning methods and Parkinson's diagnosis. However, this experiment uses brain maps of Parkinson's patients As a basis for the diagnosis of the disease, sufficient investigations have been made in the pathological diagnosis of the above Parkinson's disease, and the actual effect also reflects the experimental feasibility. This experiment was innovation.

In this experiment, a deep learning (DL) method with good effects on image recognition is used to train a large number of MRI images through a deep neural network model, learn the characteristics of the images, and then predict and diagnose the disease. The network is an optimized network based on the AlexNet network. AlexNet is a network that stood out in the ImageNet competition in 2012, and its good classification effect won the championship of the year. GoogleNet was the championship of the 2014 ImageNet competition. The basic control model used in this experiment is AlexNet and GoogleNet, and then optimize based on the AlexNet model. The optimized model achieved better results than the original model in the experiment, and also better than the classic network GoogleNet.

## 2. Deep Learning

### 2.1. Overview of Deep Learning

In 2006, deep learning was presented to people as a branch of the field of machine learning. It used multiple layers of complex structures or multiple layers composed of multiple nonlinear transformations for data processing [11]. So far, deep learning has been used in natural language processing Breakthrough progress has been made in speech recognition, especially in computer vision [12]. The advantage of deep learning is that it uses a layered and efficient feature extraction method instead of manually acquiring features, which effectively solves large-scale manual labeling work. The network hierarchy is shown in Figure 2.

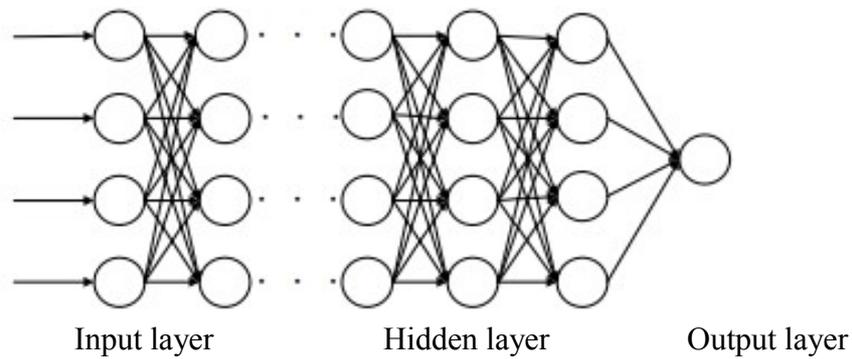

**Fig.2.** Deep neural network multilayer structure

The basic idea of deep learning is to build a multi-layer neural network as shown in the figure above, and gradually learn features from the bottom to the top to extract features. Finally, through training and learning of large amounts of data, build a corresponding network model and learn the relevant features of the training object.

## 2.2. Deep Neural Network Structure

Common deep neural networks currently include: convolutional layers, pooling layers, activation layers, etc.

In 1962, Hubel and Wiesel proposed the concept of a receptive field by studying the pupil area of the cat's eye and neurons in the cerebral cortex [13]. Later, the scholar Fukushima proposed a neural cognitive machine (neocognitron) based on this concept. The first application of the receptive field concept in the field of artificial neural networks.

The neural network containing the convolutional layer is a multi-layer neural network, which consists of multiple two-dimensional matrices, each of which is composed of multiple independent neurons. The core of the convolutional layer lies in the sharing of receptive fields and weights. The application reduces the number of parameters to be trained on the deep neural network, as shown in Figure 3.

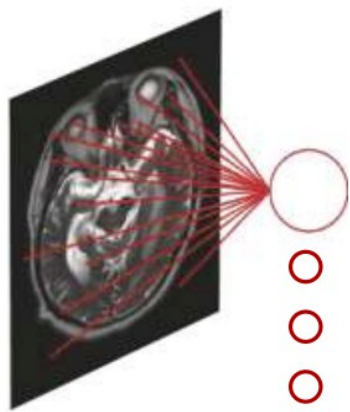

(a) Fully connected

1000 * 1000 images,
1M hidden unit
A total of 10 ^ 12 parameters

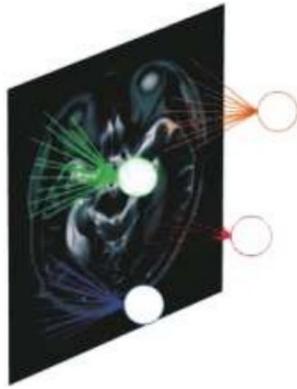

(b) Local connection

1000 * 1000 images,
1M hidden unit
Convolution kernel is 10 * 10
Total 100M parameters

**Fig.3.** Full and local connections

Weight sharing is based on the assumption that the parameters of each neuron are the same, and the receptive field is the concept of local learning of the corresponding convolution kernel. According to the comparison of the parameters in the figure above, it can be seen that the calculation is reduced by 4 orders of magnitude [14].

The pooling layer contains two types: one is average pooling, and the other is maximum pooling.

The pooling operation is a feature map reduction operation, and the main features are extracted from the original feature map.

Because the linear model has insufficient expressive power, a nonlinear model is introduced. The activation layer implements the activation of the input data, that is, a non-linear function transformation. The commonly used activation functions are Sigmoid, tanh, ReLU [15], etc., which can be based on the model effect Choose different activations, and better classification of data can be achieved through the activation layer.

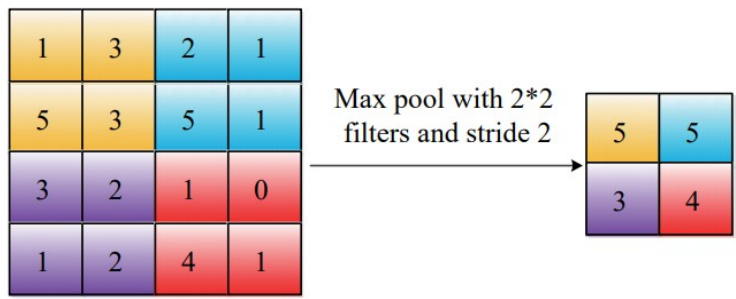

**Fig.4.** Maximum pooling

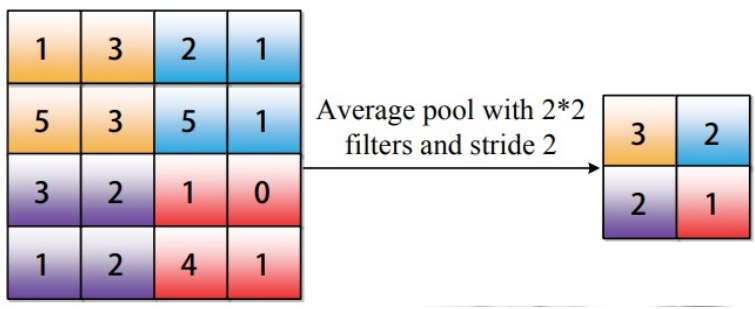

**Fig.5.** Mean pooling

## 3. Data sets and preprocessing

### 3.1. Data Set

This experimental data comes from Istanbul University Cerrahpasa Faculty of Medicine Hospital. The original data is DICOM (Digital Imaging and Communications in Medicine) images. The patient information is deleted and exported in img format by RadiAntDICOMViewer software. The details are: training set: 13 571; validation set: 2396 (15% of the training set); test set: 2237 (10% of the total data). Figure 6 shows the brain maps of PD, MSA, and Normal.

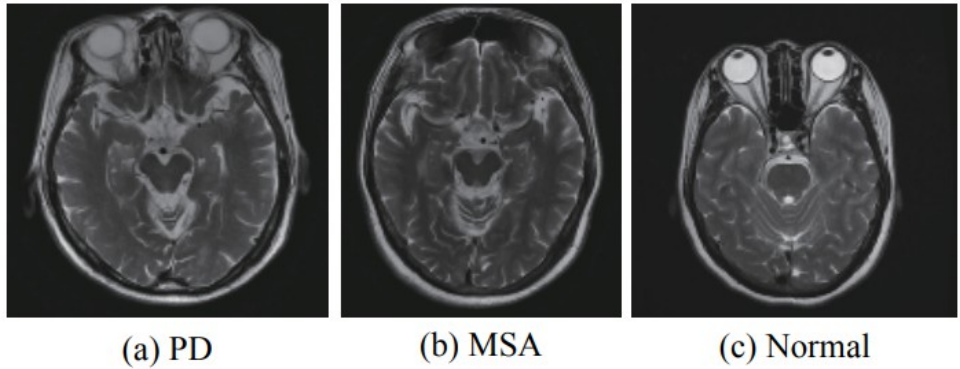

**Fig.6.** Brain maps of PD, MSA, and Normal.

### 3.2. Data Number Preprocessing

Since deep neural network training requires batch data learning features, the original data needs to be augmented during the experiment. This experiment uses the following two methods to expand the data.

### 3.2.1. Image Rotation

Image rotation refers to the process of forming a new image by rotating a certain angle around a certain point as the center, and the position of the point before and after the rotation is not changed from the center. Assume that the distance of the point ($x_0$, $y_0$) from the origin is $r$, the point The angle between the line between the origin and the $x$-axis is $b$, and the angle of rotation is $a$. The point after rotation is ($x_1$, $y_1$), then:

Location of the original point:

$$\begin{cases} x_0 = r\cos b \\ y_0 = r\sin b \end{cases}$$

Position of the point after rotation:

$$\begin{cases} x_1 = r\cos(b-a) = r\cos b \cos a + r\sin b \sin a \\ \quad = x_0 \cos a + y_0 \sin a \\ y_1 = r\sin(b-a) = r\sin b \cos a - r\cos b \sin a \\ \quad = -x_0 \sin a + y_0 \cos a \end{cases}$$

After getting the coordinates after rotation, the length and width of the rotated image will change. To recalculate the length and width of the new image, the calculation method is as follows:

Let the original image length be **srcH**, the width be **srcW**, and the center of the image as the origin. The coordinates of the upper right corner, the lower left corner, and the lower right corner are used to calculate the height and width of the rotated image, and their sizes are:

$$pLT.x = -srcW/2;\ pLT.y = srcH/2$$
$$pRT.x = -srcW/2;\ pRT.y = srcH/2$$
$$pLB.x = -srcW/2;\ pLB.y = -srcH/2$$
$$pRB.x = srcW/2;\ pRB.y = -srcH/2$$

The coordinates after rotation are set to **pLTN**, **pRTN**, **pLBN**, **pRBN**, and the sizes are:

$$pLTN.x = pLT.x * \cos a + pLT.y * \sin a$$
$$pLTN.y = -pLT.x * \sin a + pLT.y * \cos a$$
$$pRTN.x = pRT.x * \cos a + pRT.y * \sin a$$
$$pRTN.y = -pRT.x * \sin a + pRT.y * \cos a$$
$$pLBN.x = pLB.x * \cos a + pLB.y * \sin a$$
$$pLBN.y = -pLB.x * \sin a + pLB.y * \cos a$$
$$pRBN.x = pRB.x * \cos a + pRB.y * \sin a$$
$$pRBN.y = -pRB.x * \sin a + pRB.y * \cos a$$

The length and width of the rotation are set to **desHeight**, **desWidth**, and the sizes are:

$$desWidth = \max(abs(pRBN.x - pLTN.x), abs(pRTN.x - pLBN.x))$$
$$desHeight = \max(abs(pRBN.y - pLTN.y), abs(pRTN.y - pLBN.y))$$

The original image and the image after 90° rotation are shown in Figure 7.

### 3.2.2. Mirror image processing

The mirroring of the image is divided into horizontal mirroring and vertical mirroring. Let the width of the image be **width** and length be **height**, (**x, y**) are the transformed coordinates, and (**$x_0$, $y_0$**) are the coordinates of the original image.

Vertical mirror transformation:

$$\begin{cases} x = x_0 \\ y = height - y_0 - 1 \end{cases}$$

Its inverse transform:

$$\begin{cases} x_0 = x \\ y_0 = height - y - 1 \end{cases}$$

Horizontal mirroring:

$$\begin{cases} x = width - x_0 \\ y = y_0 \end{cases}$$

Its inverse transform:

$$\begin{cases} x_0 = width - x \\ y_0 = y \end{cases}$$

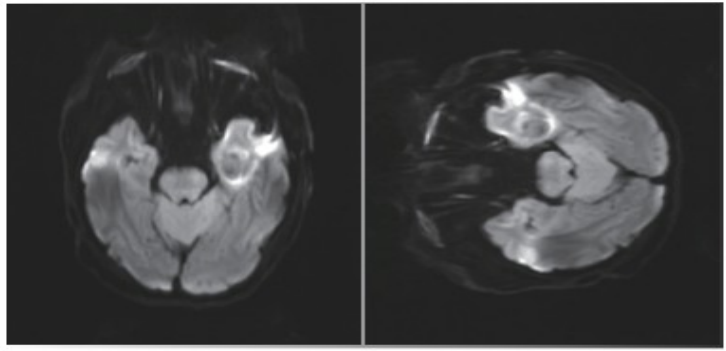

**Fig.7.** Original image (left) after 90 ° rotation (right)

In this experiment, vertical mirroring is used to achieve symmetrical exchange of left and right brain images. The comparison of vertical mirror images is shown in Figure 8.

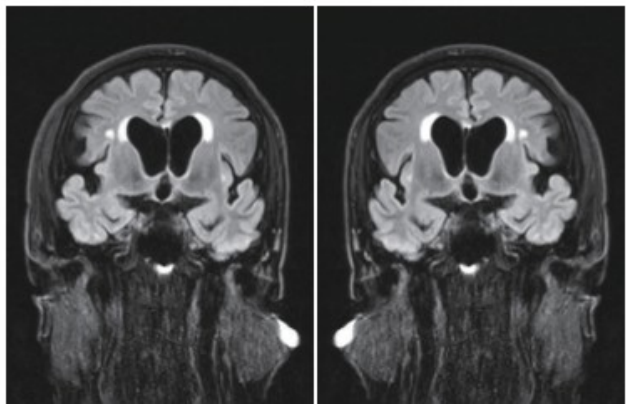

**Fig.8.** Original image (left) after mirror processing (right)

## 4. Experiments and results

### 4.1. Optimized *AlexNet* Neural Network

The *AlexNet* neural network mainly includes eight network layers, five convolutional layers, and three fully connected layers. There is a more detailed hierarchical division within each convolutional layer, as shown in Figure 9. The optimized network is shown in Figure 10. After the five-layer pooling layer, add one layer:

```
layer{
   name: "norm5"
   type: "LRN"
   bottom: "pool5"
   top: "fc6"
   lrn_param{
      local_size: 5
      alpha: 0.0001
      beta: 0.75
   }
}
```

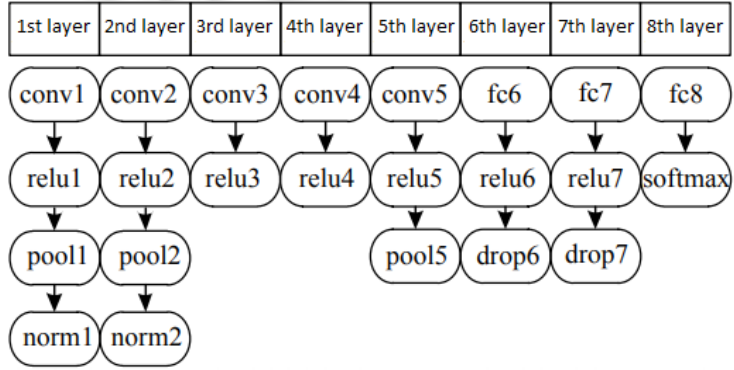

**Fig.9.** *AlexNet* neural network structure

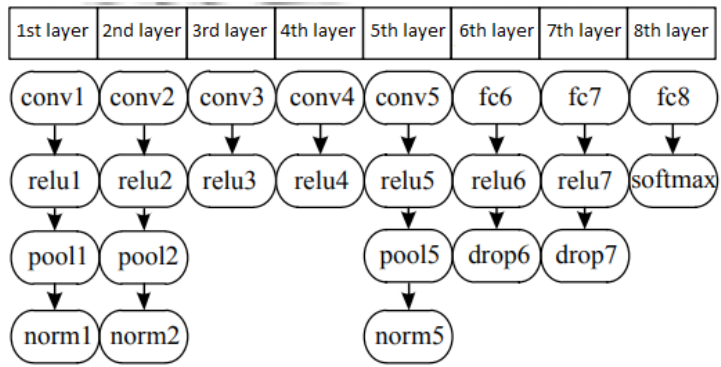

**Fig.10.** Optimized neural network structure

For example, on the network structure, it can be seen that in the fifth layer of the original *AlexNet* neural network, the norm5 network layer structure is added to this experiment. For adding norm5, the Batch Normalization operation is implemented. In the process of training a deep neural network, its complexity lies in the following The changes of the first few layers of parameters, the distribution of the input of each layer also changes during the training process, so that the learning rate during the training process needs to be set very small, which slows down the training speed. This phenomenon is called Internal covariate shift [16]. Due to the diversity of the experimental images (screenshots of different positions of the brain), a normalized layer is added by connecting the fully connected layer at the fifth layer, and the fully connected layer is input by normalization. The data is normalized as part of the model architecture, so that the model can use a higher learning rate, accelerate convergence, and improve the model effect. The algorithm is implemented as follows.

Input: Values of $x$ over a mini-batch: $B=\{x_{1...m}\}$; Parameters to be learned: $\gamma, \beta$
Output: $\{y_i = BN_{\gamma,\beta}(x_i)\}$

$\mu_\beta \leftarrow \frac{1}{m} \sum_{i=1}^{m} x_i$ // mini-batch mean

$\sigma_\beta^2 \leftarrow \frac{1}{m} \sum_{i=1}^{m} (x_i - \mu_\beta)^2$ // mini-batch variance

$\hat{x}_i \leftarrow \frac{x_i - \mu_\beta}{\sqrt{\sigma_\beta^2 + \varepsilon}}$ //normalize

$y_i \leftarrow \gamma \hat{x}_i + \beta \equiv BN_{\gamma,\beta}(x_i)$ // scale and shift

The above algorithm implements a small batch activation conversion on *x*.

In 2015, Ioffe and Szegedy applied this method to the Inception network [17] for ImageNet classification and achieved a fifth place error rate of 4.82%, which exceeds human accuracy.

***GoogleNet*** has a deep and complex network 2014 fore structure, the main innovation is limited according to the depth and width of the design, and the design of the two auxiliary Loss, is the current depth of mature neural network, as its control experiment.

### 4.2. Experimental results

Four sets of experiments were performed for the above two models, namely PD vs Normal, PD vs MSA, MSA vs Normal, and PD vs MSA vs Normal. The GPU device trained on the above model is configured with 13 NVIDIA Tesla K80, Intel (R) Xeon (R) CPU E5-2640 v4 6-core processor (2.40 GHz). 12 models of four sets of experiments can be trained simultaneously, and all models can be trained in less than 20 minutes. The experimental results are shown in Figure 11 to Figure 22.

### 4.2.1. PD vs Normal (PN)

In the classification experiments of PD and Normal, that is, Parkinson's disease and normal people (Figures 11 to 13), based on the original ***AlexNet*** experiment, the accuracy rate was improved by 0.2%, and the verification set loss was reduced by 0.01.Training The set loss achieved a reduction of 0.04.

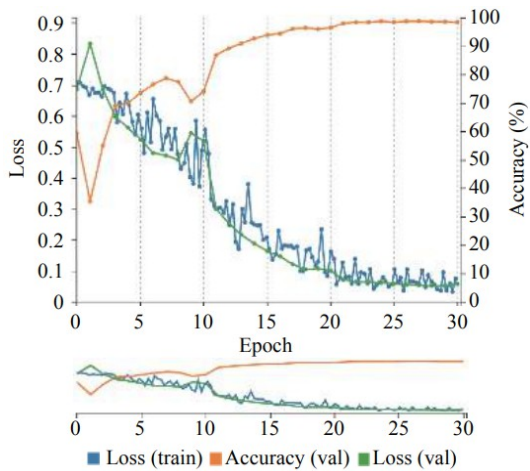

**Fig.11.** Optimized network structure

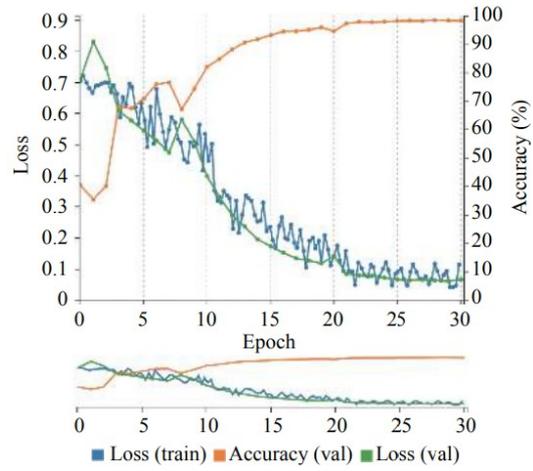

**Fig.12.** Original AlexNet network results

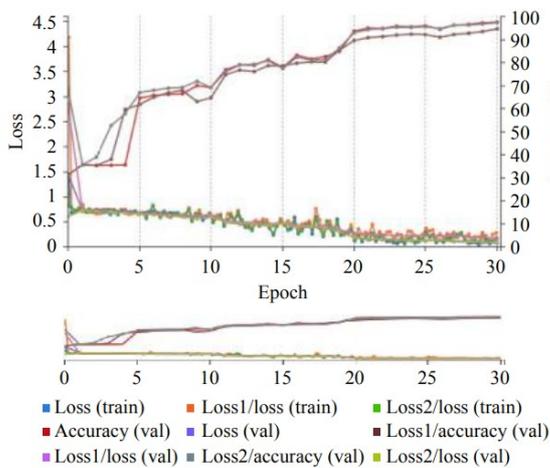

**Fig.13.** GoogleNet Network Results

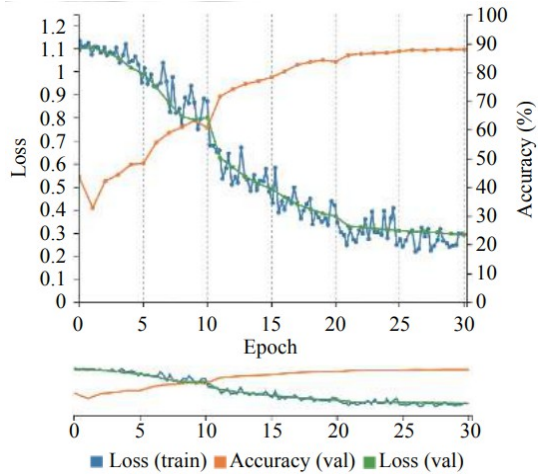

**Fig.14.** Network optimization results

## 4.2.2 PD vs MSA (PM)

In the classification experiments of PD and MSA, that is, Parkinson's disease and multisystem atrophy (Figures 14 to 16), based on the original *AlexNet* experiments, the accuracy rate has been improved by 1%, the verification set loss has remained flat, and the training set loss A reduction of 0.01 was obtained.

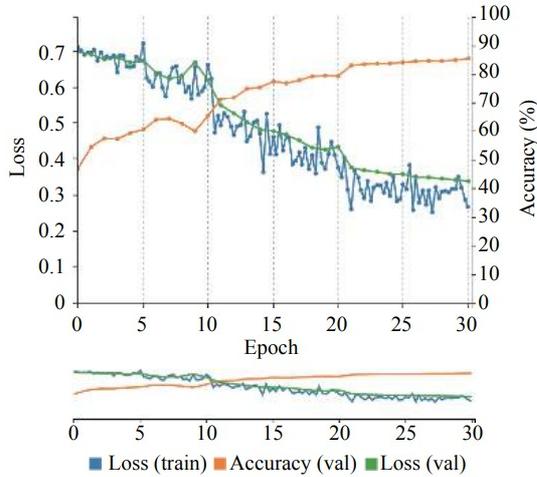

**Fig.15.** Original AlexNet network results

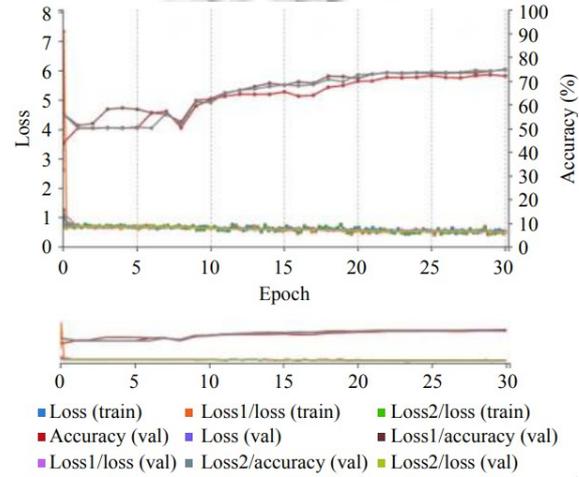

**Fig.16.** GoogleNet Network Results

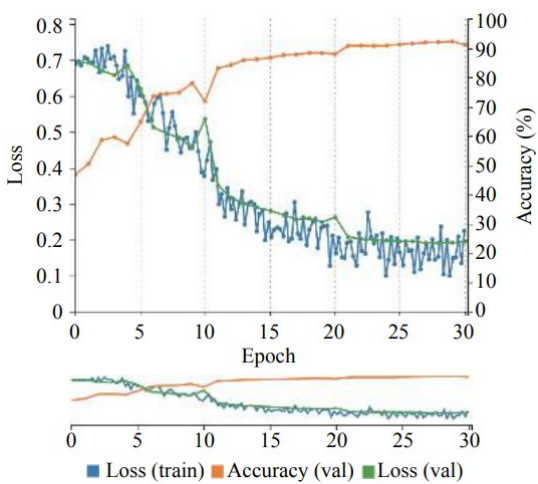

**Fig.17.** Original AlexNet network results

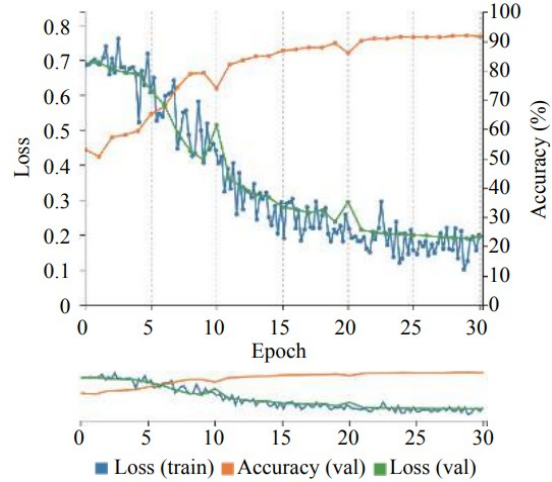

**Fig.18.** Optimized network results

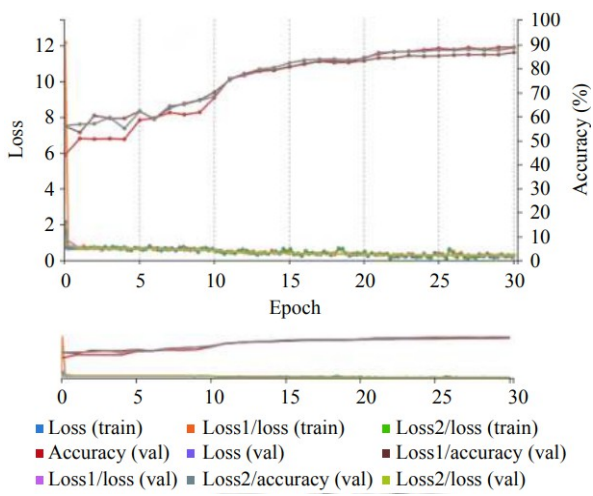

**Fig.19.** GoogleNet Network Results

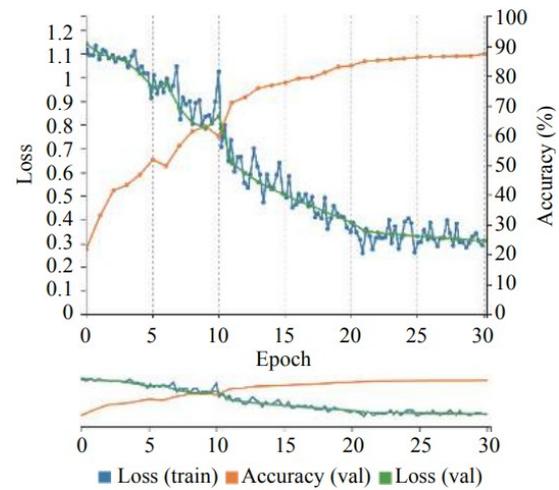

**Fig.20.** Network optimization results

### 4.2.3 MSA vs Normal (MN)

In the experiments of classification verification of MSA and Normal, that is, multi-system atrophy and normal people (Figures 17 to 19), based on the original *AlexNet* experiment, the accuracy rate was improved by 0.3%, and the verification set loss was 0.01. Reduced, training set loss was reduced by 0.03.

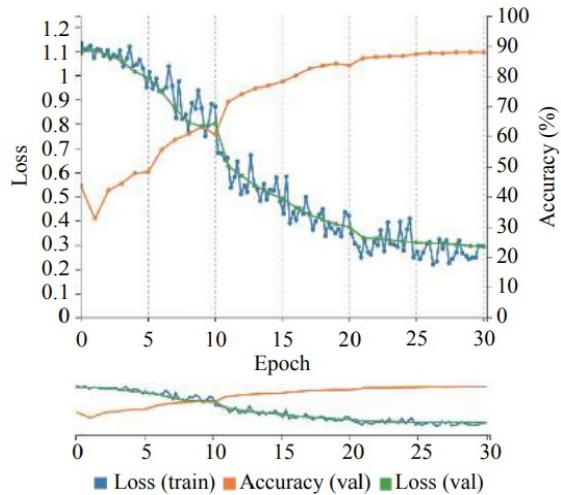 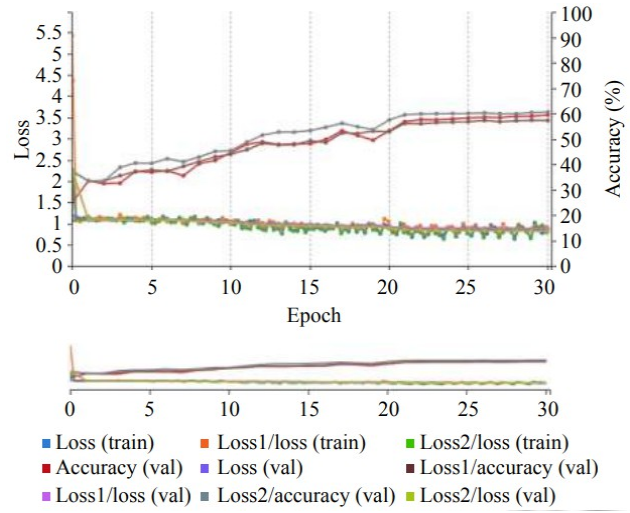

**Fig.21.** Original AlexNet network results      **Fig.22.** GoogleNet network results

### 4.2.4 PD vs MSA vs Normal (PMN)

For PD, MAS and Normal, that is, Parkinson's disease, multi-system atrophy, and normal patient classification experiments (Figures 20 to 22), the accuracy rate was improved by 0.6% based on the original *AlexNet* experiment. The validation set loss was reduced by 0.12, and the training set loss remained the same as the original experiment.

The original *AlexNet* experimental results and the optimized network experimental results and *GoogleNet* experimental results are summarized in Tables 1 to 3.

From the data analysis in the above table, the optimized network In addition to the better structure of the original *AlexNet* network, by comparing with *GoogleNet*'s Acc and Loss related data, we know that the optimized network is still better than *GoogleNet*.

|  | PN | PM | MN | PMN |
| --- | --- | --- | --- | --- |
| Acc (val) | 98.2 | 84.7 | 91.3 | 87.4 |
| Loss (val) | 0.06 | 0.33 | 0.20 | 0.31 |
| Loss (train) | 0.11 | 0.27 | 0.23 | 0.29 |

**Table 1.** Summary of the original *AlexNet* experimental indicators

|  | PN | PM | MN | PMN |
| --- | --- | --- | --- | --- |
| Acc (val) | 98.43 | 85.66 | 91.6 | 88 |
| Loss (val) | 0.05 | 0.33 | 0.19 | 0.29 |
| Loss (train) | 0.07 | 0.26 | 0.20 | 0.29 |

**Table 2.** Summary of experimental network optimization indicators

|  | PN | PM | MN | PMN |
|---|---|---|---|---|
| Acc (val) | 97.6 | 72.2 | 89 | 59.6 |
| Loss (val) | 0.07 | 0.54 | 0.26 | 0.84 |
| Loss1/Acc (val) | 94.6 | 75 | 86.8 | 57.5 |
| Loss1/loss (val) | 0.17 | 0.51 | 0.30 | 0.88 |
| Loss2/Acc (val) | 97.3 | 74.7 | 88.7 | 60.7 |
| Loss2/loss (val) | 0.09 | 0.51 | 0.27 | 0.82 |

**Table 3.** Summary of *GoogleNet* experimental indicators

## 5. Conclusion

This experiment designed an optimized version of the neural network based on the *AlexNet* neural network structure. By analyzing the experimental results of the improved *AlexNet* network and the original network in four sets of data, the improved version of *AlexNet* showed better classification results, and The optimized version of the neural network is still dominant in comparison with the experimental results of *GoogleNet*. Due to the limited amount of data in this experiment, there may be some errors, but this experiment provides reference factors for future network optimization and achieves full classification of medical images Automation, to avoid errors caused by manual screening. It also provides research significance for the early diagnosis of PD and distinguishes between PD and MSA disorders. Thanks special thanks to the experimental data and related guidance provided by Istanbul University Cerrahpasa Faculty of Medicine Hospital.

## References

1. Tsai CW, Tsai RT, Liu SP, et al. Neuroprotective effects of betulin in pharmacological and transgenic caenorhabditis elegans models of Parkinson's disease. Cell Transplantation, 2017, 26(12): 1903-1918. [doi: 10.1177/0963689717738785]

2. Scott L, Dawson VL, Dawson TM. Trumping neurodegeneration: Targeting common pathways regulated by autosomal recessive Parkinson's disease genes. Experimental Neurology, 2017, (298): 191-201. [doi: 10.1016/j.expneurol.2017.04.008]

3. 朱亨炤.大定风珠加味治疗帕金森病48例.中国医药学报, 2001, 16(6): 75. [doi: 10.3321/j.issn:1673-1727.2001.06.028]

4. Levin J, Maaß S, Schuberth M, et al. Multiple system atrophy. In: Falup-Pecurariu C, Ferreira J, Martinez-Martin P, et al., eds. Movement Disorders Curricula. Springer, Vienna. 2017. 183-192. [doi: 10.1007/978-3-7091-1628-9_17]

5. Fearnley JM, Lees AJ. Ageing and Parkinson's disease: Substantia Nigra regional selectivity. Brain, 1991, 114(5): 2283-2301. [doi: 10.1093/brain/114.5.2283]

6. Gao P, Zhou PY, Wang PQ, et al. Universality analysis of the existence of substantia nigra "swallow tail" appearance of non-Parkinson patients in 3T SWI. European Review for Medical and Pharmacological Sciences, 2016, 20(7): 1307-1314.

7. 李坤成,杨小平.帕金森病的影像学诊断.诊断学理论与实践, 2005, 4(4): 273-274. [doi: 10.3969/j.issn.1671-2870.2005.04.005]

8. Al-Fatlawi AH, Jabardi MH, Ling SH. Efficient diagnosis system for Parkinson's disease using deep belief network. Proceedings of 2016 IEEE Congress on Evolutionary Computation. Vancouver, BC, Canada. 2016. 1324-1330.

9. Hinton GE, Sejnowski TJ. Learning and relearning in Boltzmann machines. Parallel Distributed Processing: Explorations in the Microstructure of Cognition. Cambridge, MA, USA. 1986. 282-317.

10. Shamir RR, Dolber T, Noecker AM, et al. Machine learning approach to optimizing combined stimulation and medication therapies for Parkinson's disease. Brain Stimulation, 2015, 8(6): 1025-1032. [doi: 10.1016/j.brs.2015.06.003]

11. LeCun Y, Bengio YA, Hinton G. Deep learning. Nature, 2015, 521(7553): 436-444. [doi: 10.1038/nature14539]

12. Deng L, Yu D. Deep learning: Methods and applications. Foundations & Trends in Signal Processing, 2014, 7(3-4): 197-387.

13. Hubel DH, Wiesel TN. Receptive fields, binocular interaction and functional architecture in the cat's visual cortex. The Journal of Physiology, 1962, 160(1): 106-154. [doi: 10.1113/jphysiol.1962.sp006837]

14. 张巧丽,赵地,迟学斌.基于深度学习的医学影像诊断综述.计算机科学, 2017, 44(S2): 1-7.

15. Zhang C, Woodland PC. Parameterised sigmoid and ReLU hidden activation functions for DNN acoustic modelling. Sixteenth Annual Conference of the International Speech Communication Association. Dresden, Germany. 2015. 3224-3228.

16. Ioffe S, Szegedy C. Batch normalization: Accelerating deep network training by reducing internal covariate shift. Proceedings of the 32nd International Conference on Machine Learning. Lille, France. 2015. 448-456.

17. Szegedy C, Liu W, Jia YQ, et al. Going deeper with convolutions. Proceedings of 2015 IEEE Conference on Computer Vision and Pattern Recognition. Boston, MA, USA. 2015.